\documentclass[11pt]{article}
\usepackage{coling2020}
\usepackage{times}
\usepackage{url}
\usepackage{latexsym}
\usepackage{enumitem}
\usepackage{graphicx}
\usepackage{tikz}
\usepackage{caption}
\usetikzlibrary{matrix,calc}
\usepackage{multirow}
\usepackage{multicol}
\usepackage{subcaption}
\usepackage{amsmath}
\usepackage{amsfonts}

\colingfinalcopy 

\title{Identifying Depressive Symptoms from Tweets: Figurative Language Enabled Multitask Learning Framework\\
\normalsize{\color{red}{WARNING: This paper contains examples which are depressive in nature.}}}

\author{
Shweta Yadav$^{\ast}$, Jainish Chauhan$^{\dagger}$, Joy Prakash Sain$^{\ddagger}$, Krishnaprasad Thirunarayan$^{\ddagger}$ \\
\textbf{Amit Sheth}$^{\S}$, and \textbf{Jeremiah Schumm}$^{\ddagger}$ \\
  $^{\ast}$ LHNCBC, U.S. National Library of Medicine, MD, USA  \\
 $^{\dagger}$Indian Institute of Technology Gandhinagar, India \\
 $^{\ddagger}$ Wright State University, OH, USA  \\
 $^{\S}$University of South Carolina, SC, USA\\
  {\tt 
  $^{\ast}$shweta.shweta@nih.gov}, 
    {\tt 
  $^{\dagger}$chauhan.jainish@iitgn.ac.in},
      {\tt 
  $^{\S}$AMIT@sc.edu}\\
  {\tt $^{\ddagger}$\{sain.9,t.k.prasad,jeremiah.schumm\}@wright.edu
  }
  }
\date{}

\begin{document}
\maketitle
\begin{abstract}
Existing studies on using social media for deriving mental health status of users focus on the depression detection task. However, for case management and referral to psychiatrists, healthcare workers require practical and scalable depressive disorder screening and triage system.
This study aims to design and evaluate a decision support system (DSS) to \textit{reliably} determine the depressive triage level by capturing \textit{fine-grained depressive symptoms} expressed in user tweets through the emulation of Patient Health Questionnaire-9 \texttt{(PHQ-9)} that is routinely used in clinical practice.
The reliable detection of depressive symptoms from tweets is challenging because the 280-character limit on tweets incentivizes the use of creative artifacts in the utterances and figurative usage contributes to effective expression.  
We propose a novel BERT based robust multi-task learning framework to accurately identify the depressive symptoms using the auxiliary task of figurative usage detection. Specifically, our proposed novel task sharing mechanism, {\it co-task aware attention\/}, enables automatic selection of optimal information across the BERT layers and tasks by soft-sharing of parameters.
Our results show that modeling figurative usage can demonstrably improve the model's robustness and reliability for distinguishing the depression symptoms. 

\end{abstract}

\blfootnote{
    
    %
    %
    %
    This work is licensed under a Creative Commons 
    Attribution 4.0 International License.
    License details:
    \url{http://creativecommons.org/licenses/by/4.0/}.
}
\section{Introduction}
The recent survey conducted by WHO shows that a total $322$ million people in the world are living with depression. 
At its most severe, depression can lead to suicide and is responsible for $850,000$ deaths every year  \cite{world2017depression}. Early detection and appropriate treatment can encourage remission and prevent relapse \cite{halfin2007depression}. However, the stigma coupled with the depression makes patients reluctant to seek support or provide truthful answers to physicians \cite{haselton2015evolution}.
 Additionally, clinical diagnosis is dependent on the self-reports of the patient’s behavior, which requires them to reflect and recall from the past, that may have obscured over time. In contrast, social media offers unique platform for people to share their experiences in the moment, express emotions and stress in their raw intensity, and seek social and emotional support for resilience. As such, the depression studies based on social media offer unique advantages over scheduled surveys or interviews \cite{coppersmith2014quantifying,de2014mental,manikonda2017modeling,de2016discovering}. Social media self-narratives contain large amounts of implicit and reliable information expressed in real-time, that are  essential for practitioners to glean and understand user’s behavior outside of the controlled clinical environment.
Majority of these existing studies have formulated the social media depression detection task as a binary classification problem (i.e., depressive/non-depressive) and therefore are limited to only identifying the depressive users. \\
\indent To assist healthcare professionals (HPs) intervene in a  timely manner such as with an automatic triaging, it is necessary to develop an intelligent decision support system that provides HPs fine-grained depression related symptoms. The triage process is a critical step in giving care to the patients because, by prioritizing patients at different triage levels based on the severity of their clinical condition, one can enhance the utilization of healthcare facilities and the efficacy of healthcare interventions. There have been a few efforts to create datasets for capturing depression severity, however they are limited to \textbf{(1)} only clinical interviews \cite{valstar2013avec,ringeval2019avec,gratch2014distress} and questionnaires \cite{de2013predicting}, and \textbf{(2)} individuals who voluntarily participate in the study \cite{de2014characterizing}.  \\
\indent In this work, we exploit the Twitter data to identify the indications (specifically, \texttt{PHQ-9} guided symptoms) of depression. We developed a high quality dataset consisting of total $12,000$ tweets, with $3738$ tweets posted by $205$ self-reported depressed users over $2$ weeks time, which were manually annotated using \texttt{PHQ-9} questionnaire \cite{kroenke2002phq} based symptoms categories. In Table-\ref{Fig1}, we provide sample tweets associated with the nine item \texttt{PHQ-9} depression symptoms.
Our research hypothesis is that depressed individuals discuss their symptoms on Twitter that can be tracked reliably.
\begin{table*}[]
\centering
\resizebox{0.8\textwidth}{!}{%
\begin{tabular}{l|l}
\hline
\textbf{Symptom} & \multicolumn{1}{c}{\textbf{Sample Tweet}} \\ \hline
\hline
\textbf{S1: Lack of Interest} & \textit{I don't think I care about anything at all lol it's f*** w my brain , boutta go nuts} \\ 
\textbf{S2: Feeling Down} & \textit{im alone at home with no money and depressed as f***`} \\ 
\textbf{S3: Sleeping Disorder} & \textit{This is not a good night at all . Rough .} \\ 
\textbf{S4: Lack of Energy} & \textit{i have not moved all day . still in bed .} \\ 
\textbf{S5: Eating Disorder} & \textit{its good not to eat..} \\ 
\textbf{S6: Low Self-Esteem} & \textit{i am so ugly but will never stop posting pics 4 validation lol} \\ 
\textbf{S7: Concentration Problem} & \textit{my mind is screaming so many things} \\ 
\textbf{S8: Hyper/Lower Activity} & \textit{wish i didn't sit around every day wishing all my days away .why .} \\ 
\textbf{S9: Self-Harm} & \textit{Cut all my elbow up but can't feel it} \\ 
\hline
\end{tabular}
}
\caption{Sample tweets (rephrased) and their associated PHQ 9 symptoms.}
\label{Fig1}
\end{table*}
Nonetheless, user social-media post offer unique challenges as discussed below:
\begin{itemize}[noitemsep]
    \item \textbf{Usage of the figurative language:} First, the depressive users often tend to use figurative language (\textit{`FL'}) elements such as sarcasm and metaphor, to describe their symptoms. For example, one user wrote metaphorically, \textit{``My skin is paper, razor is the pen"}, while another user wrote \textit{``I want to cut myself"}. While both of these utterances refer to one specific medical concept ``Self-Harm", the first sentence utilizes paper and pen metaphorically to convey self-harm. Furthermore, previous studies utilizing social media data reported prediction errors when drug or symptom names were utilized in a figurative sense \cite{iyer2019figurative}. 
    \item \textbf{Usage of implicit sense:}   The creative expressions used by depressive users also possess implicit sense not evident from a literal reading. For example, one user expresses their desperation as, \textit{``What if life comes after death, grab my knife and find out myself."} implicitly referring  to ``Self-Harm", while another user gives a compliment with \textit{``You have a killer look."}, or captures anger through \textit{``If looks could kill, I would be dead by now."}. Other challenges include recognizing misspelled words, slangs, acronyms and unconventional contractions.
    \item \textbf{Usage of highly polysemous words:} The vocabulary of social media language offers polysemous words that require understanding of the context to determine the semantic labels. For example, ``\textit{woke up and nose started bleeding}" and ``\textit{I wish I had the nerve to press the blade deeper into my skin so I don't stop bleeding this time.}", use ``bleeding" in different contexts and senses.
\end{itemize}
 To account for this creative linguistic device widely observed in utterances of depressive users, we propose a \textbf{Fi}gurative \textbf{La}nguage enabled \textbf{M}ulti-\textbf{T}ask \textbf{L}earning framework (\texttt{FiLaMTL}) that works on the concept of task sharing mechanism \cite{ruder2017overview,yadav2018multi,yadav2019unified}.
In this work, we improve the performance and robustness of the \texttt{FiLaMTL} for the primary task of \textit{`symptom identification'} combined with the supervisory task \textit{`figurative usage detection'} in a multi-task learning setting.  We introduce a mechanism  named \textit{`co-task aware attention'} which enables the layer-specific soft sharing of the parameters for the tasks of interest. The proposed attention mechanism is parameterized with the task-specific scaling factor for BERT \cite{devlin-etal-2019-bert} layers. BERT enables even the low-resource tasks to benefit from deep bi-directional architectures and the unsupervised training framework to obtain the context-aware encoded representation. The virtue of this model is its ability to learn the task-specific representation of the input tweet by coordinating among the layers and between the tasks. \\
\textbf{Contributions:}
\begin{enumerate}[noitemsep]
    \item We propose a robust multi-task learning framework for identifying fine-grained \texttt{PHQ-9} defined symptoms from depressive tweets that takes into consideration the figurative language wired in the communication of depressive users. To the best of our knowledge, this is the first study in the depression domain that accounts for figurative language in the users social-media post.
    \item We introduce an effective way to \textit{fine-tune the BERT model} for multi-task learning using `\textit{co-task aware attention}' to better encode the feature across the different BERT layers and tasks. This mechanism allows the model to learn the layer and task-specific parameters, to control the information flow from each BERT layer, in end-to-end model training.
    \item To evaluate our study, we created a corpus of $12,155$ tweets, with $3738$ tweets annotated with 9 \texttt{PHQ-9} symptom classes  (validated by the collaborator psychiatrists). Additionally, these tweets were also labeled with the figurative classes: \textit{metaphor} and \textit{sarcasm}.  
\end{enumerate}

\section{Related Works}
Depending upon the data modalities and depressive markers, we categorize the existing literature as follows:
\begin{enumerate}[noitemsep]
    \item \textbf{Linguistic Marker:} 
    Language often reflects how people think and is a well known tool used by psychiatrists to assess the mental health condition of the people \cite{fine2006language}.
    Numerous research \cite{coppersmith2014quantifying,de2013predicting,de2014characterizing,yadav2020assessing} has shown that modeling of word-use and social language combined with network analysis has been effective in recognizing depression. 
    A widely adopted resource for understanding the linguistic patterns in mental health is the well-known Linguistic Inquiry Word Count (LIWC) \cite{pennebaker2007linguistic}. Other researchers exploited sentiment analysis \cite{xue2014detecting,huang2014detecting,yadav2018medical}, topic modeling \cite{resnik2015beyond} and emotion features \cite{chen2018mood,aragon2019detecting} to detect depression. Furthermore, substantial progress has been made with the introduction of a shared task \cite{coppersmith2015clpsych,milne2016clpsych}.
Recently, most of the existing studies \cite{yates2017depression,benton2017multi} have drifted from the traditional linguistic indicators to automated feature generation using the neural network based technique to predict or assess at-risk depressive users. 
    \item \textbf{Visual Marker:} Visual information such as head pose, body movement, facial expressions, gestures and eye blinks provide important cues in analyzing depression. 
  \newcite{girard2014nonverbal} examined if there exists a relationship between non-verbal cues and depression severity using Facial Action Coding System \cite{ekman1978facial}. 
  In another prominent study utilizing FACS, \newcite{scherer2013automatic} identified that a more downward gaze angle, dull smiles, shorter average lengths of smile, longer self-touches may predict depression.
  Several studies \cite{de2019combining,cummins2013diagnosis} have also investigated the Space-Time Interest Point (STIP) features that capture spatio-temporal changes such as facial motion and the movement in the hand, foot, and head. 
\item \textbf{Speech Marker:} Recent studies have shown the potential for exploiting speech for depression detection and monitoring \cite{cummins2015review,cummins2015analysis,scherer2014automatic}. 
Numerous research \cite{mundt2012vocal,honig2014automatic} have revealed the strength of prosodic markers, specifically the \textit{speech-rate} to analyze the level of depression. 
\newcite{moore2007critical} proposed a depression classification system based on a wide range of acoustic feature like prosodic, spectral, voice quality, and glottal feature. 
Other prominent studies \cite{mundt2012vocal,cummins2011investigation} have explored spectral features like prosodic timing measures, mel-frequency cepstral coefficients (MFCC) and glottal features to accurately classify depressed and control groups.  
    \item \textbf{Multi-modal Marker:} In recent times, there is visible surge in investigating multi-modal indicators to diagnose depression, particularly due to publicly available datasets made available through research challenges like Audio/Visual Emotion Recognition (AVEC) Workshop Depression Sub-challenge (2013-2019) \cite{schuller2011avec,valstar2013avec,ringeval2019avec} and popular Distress Analysis Interview Corpus (DAIC) \cite{gratch2014distress}. Several computational models \cite{tzirakis2017end,ringeval2017avec,ringeval2018avec} based on  machine learning and sophisticated deep learning techniques have been proposed to address the challenges posed by AVEC each year. The best system at AVEC 2019 \cite{ray2019multi} proposed an attention based fusion technique to judiciously select the feature representation obtained from multimodal source.
\end{enumerate}

\section{Corpus Creation and Analysis: D2S}
In this section, we describe how we crawled our dataset \textbf{D}epression to (\textbf{2}) \textbf{S}ymptoms (named as D2S) using the Twitter streaming application programming interface, filtered out irrelevant profiles, annotated the tweets of depressive users and verified the annotations by a psychiatrist to prepare the gold standard dataset.
\begin{enumerate}
    \item  \textbf{Dataset Crawling:} We utilized the lexicon developed by \cite{yazdavar2017semi} in collaboration with a psychologist. The lexicon contains around $1000$ depression-related terms categorized into nine categories of symptoms from \texttt{PHQ-9}. A subset of highly informative depression indicative terms from the lexicon, that are likely to be used by depressive individuals, was used as seed terms to crawl the public profiles of twitter users with at least one of those filtered terms in their profile description. Through this process, we collected $5,000$ users and their tweets.
     \item
    \textbf{Filtering and Identifying Depressed Users:} As users on social media often use sarcasm and metaphor to implicitly express their feelings, contemporary approaches that do not capture context well, miss sub-population of depressive users. To improve upon these approaches, we proceed as follows. After filtering out the retweets, we removed the profiles with less than $100$ tweets and obtained $1567$ users. To emulate
    the \texttt{PHQ-9} 
    using social media, we chunked the tweets of each profile into two week buckets. To ensure the high quality data and identify potential depressive profiles with severity level mild to severe based on \texttt{PHQ-9} scoring, we filtered the profiles based on their frequency of post. After filtering out the profiles who had not tweeted at least $5$ days in the most recent bucket, we obtained $575$ profiles.
    Although these profiles had depression-related terms in the description, due to the lack of context-sensitivity in the profile identification process, a subset of those were false positives, i.e., non-depressive. A few of these non-depressive profiles were meant to share motivational quotes for depressive users. We strictly examined the visual (i.e., profile image and shared images) and linguistic markers (i.e., profile name, description and tweets) of each of those profiles and removed the users having no depressive tweets. Finally, we obtained $205$ depressive users and selected the bucket of the most recent tweets over two weeks for annotation, which sums to $3738$ tweets. 
 \item   \textbf{Anonymization, Annotation and Verification:} Prior to annotation, we anonymized the user profiles with random numbers and replaced the mentions and URLs in tweets with strings ‘@USER’ and ‘URL’ respectively. Four native English speakers from multiple disciplines were assigned to independently annotate the tweets into $9$ categories of \texttt{PHQ-9}. The annotators were also asked to identify the tweets having usage of \textit{FL} such as sarcasm and metaphor. The annotators were provided with the definitions and samples of annotated tweets from each of those $9$ categories of \texttt{PHQ-9} and as well as \textit{FL}. The average inter-annotator reliability scores for the symptoms, depressive vs. non-depressive, and figurative classes were K=$0.83$, K=$0.87$, and K=$0.79$, respectively, based on Cohen's Kappa statistics. We resolved the conflicting annotations with the majority voting strategy and resolved the ones voted evenly by a psychiatrist. After preparing the final gold standard data, we randomly selected $100$ annotated tweets from each of the symptom categories, including the non-depressive ones and verified by a psychiatrist.
    
    \begin{table}[h]
    \centering
    \resizebox{0.8\columnwidth}{!}{
    \begin{tabular}{ccccccccccc|ccc|c}
    \hline
    \multicolumn{14}{c|}{\textbf{Depressive}} & \multirow{2}{*}{\textbf{Non-depressive}} \\
    \cline{1-14}
    \multicolumn{11}{c|}{\textbf{Symptoms}} & \multicolumn{3}{c|}{\textbf{Figurative Language}}\\
    \hline
     & \textbf{S1} & \textbf{S2} & \textbf{S3} & \textbf{S4} & \textbf{S5} & \textbf{S6} & \textbf{S7} & \textbf{S8} & \textbf{S9} & \textbf{Total} & \textbf{Sarc}  & \textbf{Meta} & \textbf{Total} & \textbf{Total} \\ 
    \textbf{\#Tweets} & 494 & 657& 261 & 301 & 473 & 1054 & 136 & 122 & 970 & 3738 & 668 & 1106 & 1485 & 8417\\ 
    \textbf{Avg.Len} & 13.02 & 14.13 & 13.03 & 13.08 & 16.08 & 12.83 & 13.31 & 15.63 & 12.33 & 12.98 & 13.17 & 13.41 & 13.24 & 10.96\\ \hline
    \end{tabular}
    }
    \vspace{-0.1cm}
    \caption{The statistics of depressive and non-depressive tweets. S[1-9]: \texttt{PHQ-9} depressive symptoms, Sarc: Sarcasm, Meta: Metaphor, \text{\#}Tweets: No. of tweets per class, Avg.Len: Average length of tweets.}
    \vspace{-0.3cm}
    \label{tbl:datasetStat}
    \end{table}
     \item \textbf{Data Analysis:} The final data\footnote{Limitation:(i) Only a sub-population (i.e., those with self-reported diagnosis) is identified by this method , (ii) Twitter users cannot be reflective of the entire population, and (iii) it cannot be verified if self-reported depressed users are being truthful} comprises 3738 depressive tweets, each tagged with the number of symptoms (single or multiple), and the presence of figurative expressions (either or both) it exhibits, and 8417 non-depressive tweets. Out of these depressive tweets, 1485 ($\sim40\%$) tweets use \textit{FL}. 
=
    We performed topic analysis to examine the usage of utterances associated with each symptom. 
    Table \ref{tbl:topicModeling} illustrates the topic distribution of each symptom. We observe from the table, to express their feelings, the depressive individuals use metaphoric phrases such as \textit{\lq{body is begging}\rq}, and \textit{\lq{feel like trash}\rq}; sarcastic expressions such as \textit{\lq{am eating ?}\rq}; implicit utterances such as \textit{\lq{up all night}\rq}, and \textit{\lq{feel myself falling}\rq}.
    \item
    \textbf{Ethical Concerns:} Psychiatric research using social media data poses several ethical concerns regarding user privacy, which should necessarily be taken into consideration \cite{hovy2016social,valdez2019ethics}. Following the ethical practices, as adopted by the previous research on Twitter data \cite{coppersmith2015clpsych}, we constructed our dataset using only public twitter profiles. We anonymized the profiles before presenting it to the annotators who pledged not to make attempts to contact or deanonymize any of the users or share the data with others. The dataset will be shared with researchers who agree to follow the similar ethical guidelines.
    \end{enumerate}
    \begin{table*}[]
    \centering
    \resizebox{0.7\linewidth}{!}{%
    \begin{tabular}{l|l}
    \hline
    \textbf{Symptom} & \multicolumn{1}{|c}{\textbf{Topics of Interest}} \\ \hline
    \textbf{S1} & \textit{bored, disgusting, sick of, tired of it, dont want to, so f** miserable, tired of being} \\ 
    \textbf{S2} & \textit{depressed, alone, isolate, given up, no friend, cant deal, want to talk, in my room} \\
    \textbf{S3} & \textit{awake, sleepless, nightmares, insomnia, cant sleep, wish sleep, up all night, body is begging} \\ 
    \textbf{S4} & \textit{exhausted, tired, weak, my energy, dont have energy, tired to look, feel myself falling} \\ 
    \textbf{S5} & \textit{binge, fasting, eating disorder, eat again, always eating, forced to eat, am eating ?} \\ 
    \textbf{S6} & \textit{failure, ugly, worthless, hate myself, fat piece, self hatred, piece of sh**, feel like trash} \\ 
    \textbf{S7} & \textit{thoughts, confused, overthinking, brain, my head, am losing, losing mind, my mind off} \\ 
    \textbf{S8} & \textit{quiet, slowly, attention, nervous, social anxiety, dead quiet,  dont wanna move} \\ 
    \textbf{S9} & \textit{cut, hang, blade, die, suicidal, rip skin, suicide attempt, car hit, kill myself, of the road} \\ 
    \hline
    \end{tabular}
    }
    \vspace{-0.1cm}
    \caption{Sample of Topics identified from depressive tweets. S[1-9]: \texttt{PHQ-9} symptoms}
    \vspace{-0.3cm}
    \label{tbl:topicModeling}
    \end{table*}

\section{Methods}
Our proposed approach to identify the depressive symptoms, is assisted by the Bidirectional Representation from Transformers (BERT) and multi-task learning \cite{yadav2020relation} with soft-parameter sharing. 
This section describes the proposed methodology for identifying the depression symptoms from user tweets. 
\subsection{Problem Definition}
Given an input tweet sequence $T$ consisting of $n$ words, i.e., $T = \{w_1, w_2 \ldots w_n\}$, our multi-label classification task is to learn the function $fun(.)$ that predicts the set of  probable classes $\bar{y_{1}}, \bar{y_{2}}, \ldots, \bar{y_{k}}$ from the set of class labels, $Y$.  Mathematically, 
\begin{equation}
    \bar{y_{1}}, \bar{y_{2}}, \ldots, \bar{y_{k}}=fun(T, \theta_1, \theta_2, \ldots, \theta_P)
\end{equation}
where, $\theta_i, (i=1, \ldots, P)$ is the model parameter. The function $fun(.)$ returns the probability of each symptom class assigned to the tweet. We choose the set of best probable class based on the particular threshold value, a hyper-parameter.
In our proposed multi-task learning framework, the primary task is symptom identification with nine labels from \texttt{PHQ-9}. We consider the figurative usage detection as the auxiliary task having three class labels: `\textit{metaphor}',`\textit{sarcasm}', and `\textit{others}'.

\subsection{Background}
BERT is one of the powerful language representation models that has the ability to make predictions that go beyond the natural sentence boundaries \cite{lin2019bert}.
Unlike CNN/LSTM model, language models benefit from the abundant knowledge from pre-training using self-supervision and have strong feature extraction capability.
It uses word-piece tokenizer \cite{wu2016google} to tokenize the input sentence. When the model uses word-piece token and randomly mask a portion of the word to predict in the masked language model (MLM) task then the model attempts to recover a piece of the word rather than the whole word. To mitigate this issue, recently, \newcite{devlin-etal-2019-bert} released an updated version of BERT, which is called Whole Word Masking (\texttt{wwm}). 
We use the pre-trained \texttt{wwm} BERT model\footnote{\url{https://bit.ly/3eQAZSY}} having $24$ Transformer layers ($L$), each having $16$ heads for self-attention and hidden dimension of $1024$. The input to the BERT model is the tweet $T=\{w_1, w_2, \ldots, w_n\}$. It returns the hidden state representation of each input word from each Transformer layer. Formally,
\begin{equation}
\label{bert_encoder}
\footnotesize
\begin{split}
    T_1, T_2, \ldots, T_L&= BERT([w_1, w_2, \ldots, w_n])\\
\text{where} \quad s_1^{i}, s_2^{i}, \ldots, s_n^{i} &= T_i \quad \text{and} \quad 
h_1, h_2, \ldots, h_L = s_1^{i}, s_1^{2}, \ldots, s_1^{L}
    \end{split}
\end{equation}
where, $s_i^{j}$ is the $i^{th}$ token representation obtained from $j^{th}$ transformer encoder, and  $h_i\in \mathbb{R}^{d}$ and $d$ is the dimension of the \texttt{[CLS]} token hidden state representation obtained from BERT. 
\subsection{\textbf{Fi}gurative \textbf{La}nguage enabled \textbf{M}ulti-\textbf{t}ask \textbf{L}earning (\texttt{FiLaMTL}) Framework}
We explore the utility of learning two tasks together in a \texttt{FiLaMTL} framework. For depression symptom identification (SI) task, \texttt{FiLaMTL} helps achieve inductive transfer from figurative usage detection (FUD) task by leveraging additional sources of information to improve performance on the primary task. We focus on designing the soft-parameter sharing rather than hard-parameter as it offers a way to effectively share the required parameters between the tasks \cite{misra2016cross}. We achieve this with \textit{co-task aware attention} module that finds the best shared representation for multi-task learning. Specifically, the proposed network models shared representations using linear combinations, and learns the optimal combinations for the primary and the auxiliary tasks. 
Let us denote the hidden states (from eq. \ref{bert_encoder}) for primary task (SI), , as $H_s \in \mathbb{R}^{L \times d}$ and the auxiliary task (FUD), as $H_f \in \mathbb{R}^{L \times d}$. For a given layer $l \in L$, we compute the effective shared representation as follows:
\begin{equation}
\label{attention}
    \begin{split}
    r_s^l &= \alpha_{(s, s)} \times \beta_{(l, s, s)} \times h_s^{l} + \alpha_{(s, f)} \times \beta_{(l, s, f)} \times h_f^{l} \\
    r_f^l &= \alpha_{(f, f)} \times \beta_{(l, f, f)} \times h_s^{l} + \alpha_{(f, s)} \times \beta_{(l, f, s)} \times h_f^{l} \\
    \end{split}
\end{equation}
where $h_s^{l} \in \mathbb{R}^{d}$ and  $h_f^{l} \in \mathbb{R}^{d}$ are the hidden state representations obtained from $l^{th}$ BERT layer for SI and FUD respectively.  $r_s^{l} \in \mathbb{R}^{d}$ and  $r_f^{l} \in \mathbb{R}^{d}$ are modified hidden state representation of $l^{th}$ BERT layer, after applying the effective soft-sharing of features across the two tasks. We will discuss the scaling factors $\alpha$ and $\beta$ shortly.
\subsubsection{Soft-parameter Sharing between Tasks} In multi-task learning, inductive bias of auxiliary task helps to improve the performance of primary task. However, the parameter sharing between the tasks is non-trivial, as we need an optimal strategy to improve the performance of primary task. Towards this end, we devise a strategy to automatically learn the factor by which a feature from a particular task need to be accommodated for learning the optimal set of shared features for a given task. This co-task aware sharing of the features leads to the optimal linear combination of feature spaces across the task. Given the two tasks: ``\textit{symptom identification}'' and ``\textit{figurative usage detection}'', we learn how much of each task's features contribute to form the shared feature space, which leads to the overall improvement of the tasks. We achieve this by introducing a ``\textit{co-task factor matrix}'' $\alpha \in \mathbb{R}^{T \times T} $, where $T$ is the number of tasks at hand. In our case $T=2$, as we are dealing with two tasks here.  An element $\alpha_{(x, y)}$ of the matrix $\alpha$ denotes  ``\textit{the factor of which $y^{th}$ task feature obtained from a particular layer of BERT should contribute to the shared-feature representation for $x^{th}$ task}''. Moreover, this matrix is learned by end-to-end training of the proposed multi-task learning framework.

\begin{minipage}{.4\textwidth}
\begin{tikzpicture}
\def\xs{1} 
\def\ys{0.5} 
\def\nm{3} 
\foreach \x in {\nm,...,1}
{

\matrix [draw, 
         fill=white, 
         ampersand replacement=\&] 
(mm\x)
at(-\x * \xs, -\x * \ys) 
{
    \node {$\alpha_{1, \x}$};\\
    \node {$\alpha_{2, \x}$};\\
    \node {$\alpha_{3, \x}$};\\
};

}
\draw [dashed,gray](mm1.north west) -- (mm\nm.north west);
\draw [dashed,gray](mm1.north east) -- (mm\nm.north east);
\draw [dashed,gray](mm1.south east) -- (mm\nm.south east);
\end{tikzpicture}
\captionsetup{width=\linewidth}

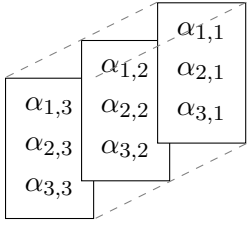
\captionof{figure}{Representation of co-task factor matrix for the three tasks.}
\label{alpha-matrix}
\end{minipage}
\hfill
\begin{minipage}{.5\textwidth}
\begin{tikzpicture}
\def\xs{1} 
\def\ys{0.5} 
\def\nm{5} 
\foreach \x in {\nm,...,1}
{

\matrix [draw, 
         fill=white, 
         ampersand replacement=\&] 
(mm\x)
at(-\x * \xs, -\x * \ys) 
{
    \node {$\beta_{\x, 1, 1}$}; \& \node {$\beta_{\x, 1, 2}$};\\
    \node {$\beta_{\x, 2, 1}$}; \& \node {$\beta_{\x, 2, 2}$};\\
};
}

\draw [dashed,gray](mm1.north west) -- (mm\nm.north west);
\draw [dashed,gray](mm1.north east) -- (mm\nm.north east);
\draw [dashed,gray](mm1.south east) -- (mm\nm.south east);
\end{tikzpicture}

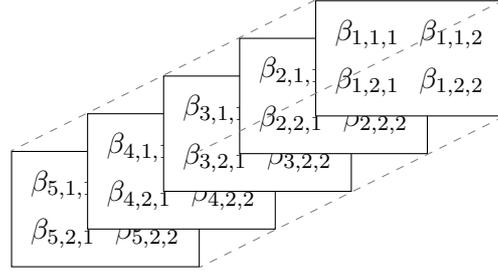
\captionof{figure}{Representation of layer factor matrix for five layers and two tasks.}
\label{beta-matrix}
\end{minipage}

\subsubsection{Soft-parameter Sharing between Layers}
Given input tokens and task $t$, BERT produces the set of layer activation $h_t^1, h_t^2, \ldots, h_t^{L}$ in the form of hidden state representation. Each layer of the BERT attempts to address certain problem \cite{tenney-etal-2019-bert} and feed the information to the upper layer. \newcite{tenney-etal-2019-bert} discovered that information learned at a few layers of BERT is sufficient to reliably model and address the lower-level tasks of an NLP pipeline such as parts-of-speech tagging, but, to model the higher level tasks such as relation extraction or co-reference resolution, we need many layers. 
The hidden state representation $h_{t_1}^{l}$, obtained from the $l^{th}$ layer of BERT for a given task $t_1$, may not be as useful for another task $t_2$. It always depends on task complexity. 
Inspired by this, we introduce the soft-parameter sharing among the BERT layers. Similar to the soft-parameter sharing between tasks, we propose a mechanism to automatically learn the factors by which a feature from a particular BERT layer needs to be accommodated for learning the optimal set of shared features for a given task. We achieve this by introducing a ``\textit{layer factor matrix}'' $\beta \in \mathbb{R}^{L \times T \times T} $, where $L$ and $T$ denote the number of BERT layers and the number of tasks respectively.  An element $\beta_{(x, y, z)}$ of the matrix $\beta$ denotes  ``\textit{the factor of which $z^{th}$ task feature obtained from $x^{th}$ layer of BERT should contribute to the shared-feature representation for $y^{th}$ task}''. Similar to the co-task factor matrix $\alpha$, the layer factor matrix $\beta$ is also a network parameter and can be learned by end-to-end training of proposed multi-task learning framework. We shown the hypothetical matrices for $\alpha$ and $\beta$ in Fig \ref{alpha-matrix} and \ref{beta-matrix} respectively.
\subsection{Symptom Identification and Figurative Usage Detection}
For each task, we obtained the effective shared-feature representation from each BERT layer. The final feature is obtained by the average pooling of each individual feature as follows:
\begin{equation}
\footnotesize
\label{final-feature}
\begin{split}
    z_s =  \frac{1}{L}\sum_{i=1}^{i=L} f(\mathbf{W_s}.r_s^i + \mathbf{b_s})  \quad \textrm{and} \quad
z_f =  \frac{1}{L}\sum_{i=1}^{i=L} f(\mathbf{W_f}.r_f^i + \mathbf{b_f}) 
\end{split}
\end{equation}
where $z_s \in \mathbb{R}^{d}$ and $z_f \in \mathbb{R}^{d}$ correspond to the final pooled features for the task symptom identification and figurative usage detection respectively. $\mathbf{W_s}$, $\mathbf{W_f}$, $\mathbf{b_s}$ and $\mathbf{b_f}$ are the weight and bias matrices and $f$ is a non-linear activation. For symptom identification, we employ a feed-forward network with sigmoid activation function to find the probability of a class label belonging to a given tweet, 
\begin{equation}
\footnotesize
\label{final-label}
\begin{split}
    l_s =   sigmoid(\mathbf{W_{sl}}. z_s + \mathbf{b_{sl}})  \quad \textrm{and} \quad
l_f =  sigmoid(\mathbf{W_{fl}}.z_f + \mathbf{b_{fl}}) 
\end{split}
\end{equation}
where $l_s$ and $l_f$ are logits for the symptom identification and figurative usage detection tasks respectively. $\mathbf{W_{sl}}$, $\mathbf{W_{fl}}$, $\mathbf{b_{sl}}$ and $\mathbf{b_{fl}}$ are the weight and bias metrices.


\section{Experiments}
We will first provide detail about baseline models and then present our results on SI and FUD task. Later, we will assess the performance of our approach on depression detection task. 
\subsection{Implementation Details}
We shuffle the D2S dataset and split it into 70\% training (TRAIN), 10\% development (DEV), and 20\% test (TEST). For both symptom identification (SI) and figurative usage detection (FUD) models, we have chosen the hyper-parameters using the development set. In all of our experiments, we have fine-tuned the BERT-wwm model for $10$ epochs with the batch size of $32$. We fine-tune and extracted the features from top three layers of the BERT model. In the proposed FiLaMTL framework, the overall loss of the network is the weighted factor of the loss computed for both the tasks. The network is trained with the binary-cross entropy loss function for  both tasks. We set the weight $0.7$ for symptom identification and $0.3$ for figurative usage detection task. We use sigmoid activation function as the non-linear activation to project the BERT hidden state representation to another representation of dimension $256$.
 We used Adam optimizer \cite{kingma2014adam} with a fixed learning rate of $0.0001$. 
For regularization, we  used dropout \cite{srivastava2014dropout} with a value of $0.5$ on each of the hidden layers. 
We then ran each best model on TEST, and report recall, precision, and F1-Score.

\subsection{Baseline Models and Results}
We compare the \texttt{FiLaMTL} with the highly competitive baseline models and evaluated the model on Precision, Recall, and weighted F1-Score. Since BERT has already demonstrated remarkable performance on multiple NLP tasks over SOTA deep learning (DL) methods, we restricted ourselves to using BERT over DL techniques as our baseline model discussed below:\\
    \textbf{(1) \textit{STL-BERT}:} This is a domain-adapted BERT-based model proposed for the SI and FUD tasks, which fine tuned the BERT model on corresponding dataset.\\
    \textbf{(2) \textit{MTL-H-BERT}, Dense:} A variation of the multi-task BERT model where a single BERT model generate the features for both the tasks. The features are transformed to another representation by the task-specific dense layer.\\
    \textbf{\textbf{(3)} \textit{MTL-S-BERT, Cross-Stitch}:} This model is the re-implementation of \newcite{misra2016cross}, where the model learns an optimal combination of shared and task-specific representations using soft-parameter sharing via cross-stitch units. \\
    \textbf{(4) \textit{MTL-S-BERT, Co-Attention}:} This model was inspired by the framework of \newcite{lu2016hierarchical}. Firstly, we compute the word-level attention weight as discussed in \newcite{lu2016hierarchical} for the hidden state representation of both the tasks. These weights were multiplied with the corresponding hidden state representation to compute the attentive features. Similar to  \textbf{\textit{MTL-S-BERT, Cross-Stitch}}, we employed the cross-stitch units to obtained the final hidden state representation for both the tasks.\\
\indent Table-\ref{RESULT} provides a comparative summary of the results of our proposed approach over the baseline models demonstrating that our `\textit{co-task-aware}' multitask \texttt{FiLaMTL} model outperforms the SOTA single task learning model and the variations of BERT inspired multi-task learning models. Basically, we train MTL model in two different ways: \textit{hard-parameter sharing} and \textit{soft-parameter sharing}. We can visualize from Table-\ref{RESULT}, the multitask learning framework based on the soft-parameter sharing (MTL-S-BERT, Cross-Stitch) can assist the performance of the main task as well as in the FUD task over the single task model. 
However, multi-task hard parameter sharing model (MTL-H-BERT, Dense) was found to be useful only in the FUD over SI task. This may be due to the noise in the dataset over the tasks, which prevents to learn task-specific efficient representation required to correctly identify the symptom from the input tweet. We also observe that soft-parameter sharing based baseline model (MTL-S-BERT, Co-Attention) could not produce the desired results, because of the additional attention mechanism over the strong self-attention mechanism, overfits the model which leads to the degradation in the performance. 
The existing strategy of the parameter-sharing shows the inconsistency in the performance of the multi-task learning framework. We exploited variation of soft-parameter sharing to further understand the relevance of \textit{co-task aware} attention in the multi-task learning setting. Our \texttt{FiLaMTL} outperforms both the hard-parameters and soft-parameters sharing based baseline models on both the tasks (Table \ref{RESULT}). This also demonstrates that providing information about \textit{FL} to the BERT model significantly improves the performance of the model and thus enabling generalization to other tasks related to text classification where there is extensive usage of \textit{FL}.
\begin{figure}[!t]
\begin{minipage}{\textwidth}
\begin{minipage}[c]{0.63\textwidth}
\resizebox{\textwidth}{!}{%
\begin{tabular}{l|ccc|ccc}
\hline
\multirow{2}{*}{\textbf{Models}} & \multicolumn{3}{c|}{\begin{tabular}[c]{@{}c@{}} \textbf{Symptom Identification}\\ (SI)\end{tabular}} & \multicolumn{3}{c}{\begin{tabular}[c]{@{}c@{}}\textbf{Figurative Usage Detection }\\ (FUD)\end{tabular}} \\ \cline{2-7} 
& {P}       & {R}       & {F1}      & {P}       & {R}       & {F1}      \\ \hline
STL-BERT    & $74.64$      &         $71.46$           & $73.02$              &  $63.32$              & $71.35$            &  $67.09$            \\ 
MTL-H-BERT, Dense    & $73.76$               &  $72.15$         &  $72.95$             & $62.97$                &  $75.17$          &  $68.53$              \\ 
MTL-S-BERT, Cross-Stitch \cite{misra2016cross}   &  $76.17$         &   $71.00$          &  $73.50$            & $68.48$     &  $77.35$           &  $72.65$               \\ 
MTL-S-BERT, Co-Attention \cite{lu2016hierarchical}    & $74.09$ 		  &  $70.43$         & $72.21$              & $69.28$                & $78.44$           & $73.58$               \\ 
\texttt{FiLaMTL}      & $\textbf{76.46}$		   &  $\textbf{73.65 }$        & $\textbf{75.03}$              &  $\textbf{75.67}$	  &  $\textbf{75.44}$          &  $\textbf{75.55}$             \\ 
\hline
\end{tabular}
}
\captionof{table}{Performance comparisons of our proposed approach with baselines for identifying depressive symptoms and figurative usage. The results are reported for only positive classes.}
\label{RESULT}
\end{minipage}
\hfill
\begin{minipage}[c]{0.35\textwidth}
\centering
    \resizebox{\linewidth}{!}{%
\begin{tabular}{l|l|l}
\hline
\textbf{Dataset} & \textbf{Model}  & \textbf{Accuracy} \\ \hline
\multirow{2}{*}{D2S} & BERT   & $96.91$ \\ 
 & \texttt{FiLaMTL}-\textit{fine-tuned}    & $97.44$ \\ \hline
\multirow{2}{*}{CLPsych} & BERT  & $70.01$ \\ 
 & \texttt{FiLaMTL}-\textit{fine-tuned}  & $70.79$ \\ \hline
\end{tabular}
}
\captionof{table}{Evaluation of \texttt{FiLaMTL} on our D2S dataset and CLPsych 2015 dataset for depression detection task.}
\label{DD}
\end{minipage}
\end{minipage}
\end{figure}
\begin{table*}[h]
\centering
\begin{tabular}{l|l|l|l}
\hline
\textbf{No.} & \textbf{Control} & \textbf{Depressed} & \textbf{Total} \\ \hline
\# Users in train &151  &69  & 220 \\ \hline
\# Users in test & 300 & 150 & 450 \\ \hline
\# Tweets in train &254786  & 110585 & 365371 \\ \hline
\# Tweets in test & 541282 & 284777 & 826059 \\ \hline
\end{tabular}
\caption{Dataset Statistics on the CLPsych 2015 shared task corpus used in our experiments for depression detection.}
\label{tab:my-table1}
\end{table*}
\paragraph{Evaluating the \texttt{FiLaMTL} on Depression Detection Task:} To further verify the effectiveness of our proposed \texttt{FiLaMTL} model, we utilized a transfer learning procedure, where an intermediate shared model obtained on the \textit{SI} and \textit{FUD} task is fine-tuned on the depression detection (DD) task. Towards that, we experimented with \textbf{(a)} STL-BERT and \textbf{(b)} \texttt{FiLaMTL}- fine-tuned for DD task, on D2S corpus and the bechmark CLPsych dataset \cite{coppersmith2015clpsych}. The data statistics can be viewed in Table-\ref{tab:my-table1}. In \texttt{FiLaMTL} (fine-tuned), we initialize the parameters of the BERT model with the obtained weighted from the \texttt{FiLaMTL} model (BERT model of FUD task) reported in Table \ref{RESULT}. Our motivation is that first fine-tuning on the \textit{FUD} and \textit{SI} task can assist the LMs to adapt to the depression domain with some understanding of figurative usage detection, thus making the training on DD more stable. Table-\ref{DD}, summarizes the results on DD task, after the transfer learning procedure. It can be noticed that fine-tuning \texttt{FiLaMTL} model improves the performance over vanilla BERT model on both the datasets \footnote{We were not able to compare \texttt{FiLaMTL} over existing model developed on CLPsych dataset, due to: (1) different dataset statistics of what we obtained from organizers and the dataset used by the participants and (2) multiple ways of evaluations. }. This proves that \texttt{FiLaMTL} can be generalized for other tasks related to biomedical NLP task where there is extensive usage of \textit{FL}.\\

\noindent \textbf{Analysis:} To get a deeper insight into how \texttt{FiLaMTL} performed over the baseline models, we examined the classification of tweets on SI task and came up with the following observations:
\begin{enumerate}
    \item 
 \textbf{Understanding figurative sense:} Table-\ref{tab:compare} shows the capability of our model to capture sarcastic and metaphoric senses in the utterances of depressive users. Our model performs better than STL-BERT in handling figurative tweets. The main reason why STL-BERT model misclassifies sarcastic or metaphoric tweets is because BERT has been trained on BookCorpus and Wikipedia corpus which has fewer \textit{FL} fragments compared to that in social media.
 \item
 \textbf{Understanding implication:} For the tweets where depression was implicit (cf., Table-\ref{tab:compare}), most of the baseline models including MTL-series were prone to misclassification. As BERT based models tend to  capture only local information available in a tweet, it fails to understand the implicit context of the subject. For example, if a tweet contains a keyword `\textit{sleep}', the model will likely classify it as belonging to the PHQ class 3  related to sleeping disorder, without necessarily determining a different contextual use (such as ``\textit{permanent land of nod}" and ``\textit{going to sleep early tonight}"). However, our model with the \textit{co-task aware attention} information sharing unit tends to have better coverage for identifying the depressive symptoms.
 \end{enumerate}
\begin{table*}[]
\resizebox{\textwidth}{!}{%
\begin{tabular}{llllllll}
\hline
\multicolumn{1}{l}{\multirow{2}{*}{\textbf{Case}}} & \multicolumn{1}{c}{\multirow{2}{*}{\textbf{Tweets}}} & \multicolumn{1}{c}{\multirow{2}{*}{\textbf{Actual Labels}}} & \multicolumn{4}{c}{\textbf{Predicted Label}} \\ \cline{4-8} 
\multicolumn{1}{l}{} & \multicolumn{1}{c}{} & \multicolumn{1}{c}{} & \multicolumn{1}{l}{\textbf{STL-BERT}} & \multicolumn{1}{l}{\textbf{MTL-H-BERT}} & \multicolumn{1}{l}{\begin{tabular}[c]{@{}c@{}}\textbf{MTL-S-BERT,} \\ \textbf{Cross-Stitch}\end{tabular}} &
\multicolumn{1}{l}{\begin{tabular}[c]{@{}c@{}}\textbf{MTL-S-BERT,} \\ \textbf{Co-attention}\end{tabular}} &
\multicolumn{1}{l}{\texttt{\textbf{FiLaMTL}}} \\ \hline
\multirow{2}{*}{\textbf{Understanding figurative sense}} & \textit{\textbf{T1: holy sh**. i look like death}} & Low Self-Esteem & \begin{tabular}[c]{@{}l@{}}Low Self-Esteem,\\ Self-Harm\end{tabular} & \begin{tabular}[c]{@{}l@{}}Low Self-Esteem,\\ Self-Harm\end{tabular} & Low Self-Esteem 
& Self-Harm
& Low Self-Esteem \\
 & \textit{\begin{tabular}[c]{@{}l@{}}\textbf{T2: hang on a rope or bated breath,}\\ \textbf{whichever you prefer}\end{tabular}} & Self Harm & Feeling Down & \begin{tabular}[c]{@{}l@{}}Feeling Down,\\ Lack of Interest\end{tabular} & Feeling Down & \begin{tabular}[c]{@{}l@{}}Self-Harm,\\ Feeling Down\end{tabular} & Self-Harm \\ \hline
\multirow{2}{*}{\textbf{Understanding implication}} & \textit{\begin{tabular}[c]{@{}l@{}}\textbf{T3: i want a zombie to read my} \\ \textbf{nutrition label, be happy to see} \\ \textbf{it say low fat , and then eat me}\end{tabular}} & Low Self-Esteem & Eating Disorder & \begin{tabular}[c]{@{}l@{}}Eating Disorder,\\ Self-Harm\end{tabular} & \begin{tabular}[c]{@{}l@{}}Eating Disorder,\\ Low Self-Esteem\end{tabular} & Eating Disorder
& Low Self-Esteem \\
 & \textit{\textbf{T4: insomia can i pls sleep forever}} & \begin{tabular}[c]{@{}l@{}}Sleeping Disorder,\\ Self-Harm\end{tabular} & Sleeping Disorder & Sleeping Disorder & Sleeping Disorder
 &\begin{tabular}[c]{@{}l@{}}Sleeping Disorder, \\ Self-Harm\end{tabular}
 &
 \begin{tabular}[c]{@{}l@{}}Sleeping Disorder, \\ Self-Harm\end{tabular} \\ \hline
\end{tabular}
}
\caption{Qualitative analysis of our proposed model, \texttt{FiLaMTL}, with the baseline models}
\label{tab:compare}
\end{table*}
\begin{figure}
\begin{minipage}{\textwidth}
 \begin{minipage}[t]{0.6\textwidth}
\centering
\resizebox{\textwidth}{!}{%
\begin{tabular}{llll}
\hline
\multirow{1}{*}{\textbf{Error Types}} & \multicolumn{1}{c}{\multirow{1}{*}{\textbf{Tweets}}} & \multicolumn{1}{c}{\multirow{1}{*}{\textbf{Actual Labels}}} & \multirow{1}{*}{\textbf{Predicted Label}} \\ \hline
\multirow{2}{*}{\textbf{Ambiguity}} & \textit{T1: sorry i was such a failure} & Low Self-Esteem & \begin{tabular}[c]{@{}l@{}}Low Self-Esteem,\\ \textcolor{red}{\textit{Feeling Down}}\end{tabular} \\ \cline{2-4} 
 & \textit{\begin{tabular}[c]{@{}l@{}}T2: its worth noting that im \\ not worth noting\end{tabular}} & Low Self-Esteem & \begin{tabular}[c]{@{}l@{}}Low Self-Esteem,\\ \textcolor{red}{\textit{Feeling Down}}\end{tabular} \\ \hline
\multirow{2}{*}{\textbf{Cryptic tweets}} & \textit{T3: shut up stomach} & Eating Disorder & \textcolor{red}{\textit{None}} \\ \cline{2-4} 
 & \textit{T4:  im done} & Self-Harm & \textcolor{red}{\textit{Feeling Down}} \\ \hline
\multirow{2}{*}{\textbf{Multiple Symptoms}} & \textit{\begin{tabular}[c]{@{}l@{}}T5: it really sucks being strong \\ all the time. its so draining and \\ when youre all depleted it feels \\ like youre underwater.\end{tabular}} & \begin{tabular}[c]{@{}l@{}}Lack of Interest,\\ Lack of Energy,\\ Concentration Problem,\\ Hyper/Lower Activity\end{tabular} & \begin{tabular}[c]{@{}l@{}}\textcolor{red}{\textit{Lack of Energy}},\\ Hyper/Lower Activity\end{tabular} \\ \cline{2-4} 
 & \textit{\begin{tabular}[c]{@{}l@{}}T6: im 24/7. online. bored.\\ hungry . sleepy\end{tabular}} & \begin{tabular}[c]{@{}l@{}}Lack of Interest,\\ Sleeping Disorder,\\ Eating Disorder\end{tabular} & \begin{tabular}[c]{@{}l@{}}\textcolor{red}{\textit{Feeling Down}},\\ \textcolor{red}{\textit{Lack of Energy}}, \\Sleeping Disorder\end{tabular} \\ \hline
\end{tabular}
}
\captionof{table}{Exemplar description showing prominent errors made by our proposed approach.}
\label{tab:2}
 \end{minipage}
 \hfill
 \begin{minipage}[t]{0.4\textwidth}
     \centering
     \resizebox{\textwidth}{!}{%
\begin{tabular}{lll}
\hline
\multicolumn{1}{c}{\textbf{Tweets}} & \multicolumn{1}{c}{\textbf{Actual Label}} & \multicolumn{1}{c}{\textbf{Predicted Label}} \\ \hline
\textit{\begin{tabular}[c]{@{}l@{}}T1: hang on a rope or bated breath,\\ whichever you prefer.\end{tabular}} & Sarcasm & Sarcasm \\ \hline
\textit{\begin{tabular}[c]{@{}l@{}}T2: If looks could kill, I would be dead \\ by now.\end{tabular}} & Sarcasm & Sarcasm \\ \hline
\textit{\begin{tabular}[c]{@{}l@{}}T3: people treat me like a god. they ignore \\ my existence until they need something \\ from me .\end{tabular}} & Metaphor & \begin{tabular}[c]{@{}l@{}}\textcolor{red}{\textit{Sarcasm}}, \\ Metaphor\end{tabular} \\ \hline
\textit{T4: i dont see a future} & Sarcasm & \textcolor{red}{\textit{Others}} \\ \hline
\end{tabular}
}
\captionof{table}{Qualitative analysis of \texttt{FiLaMTL} in identifying figurative language. }
\label{tab:QS-F}
 \end{minipage}
 \end{minipage}
 \end{figure}

\noindent \textbf{Error Analysis:} Following are the major errors made by our approach on SI task:
\begin{enumerate}[noitemsep]
    \item 
    \textbf{Ambiguity:} \texttt{PHQ-9} classes  related to sleeping disorder, eating disorder, and self-harm are easy to distinguish. However  classes such as  PHQ-1 (feeling down) and PHQ-6 (low self-esteem) are difficult  to separate because of overlapping expressions, often leading to misclassification. 
    As observed in Table-\ref{tab:2}, both these classes are semantically similar, challenging manual labelling. 
    \item
  \textbf{Cryptic tweets:} Our model is unable to handle tweets that are only a few words long. The lack of context required for robust identification of symptoms can only be remedied by consulting past user interactions and communications.
  \item
  \textbf{Multiple Symptoms:} The \texttt{FiLaMTL} is unable to predict all the \texttt{PHQ-9} classes indicated by a tweet leading to incompleteness as shown in Table-\ref{tab:2}, tweet T5 and T6. 
  \end{enumerate}
\section{Conclusion}
In this research, we explored a new dimension of social media in Twitter to identify depressive symptoms. Towards this end, we created a new benchmark dataset (\texttt{D2S}) for identifying \textit{fine-grained PHQ-9 emulated depressive symptoms} that contains figurative language. We also introduce a robust BERT based MTL framework that jointly learns to automatically discover complementary features required to identify the symptoms with the help of the auxiliary task of \textit{figurative usage detection}. Our experimental results convincingly show the effectiveness of introducing figurative usage detection for depressive symptoms identification. In future, we aim to enhance the dataset with the other modalities like image and memes to assist the model in better understanding of figurative sense in symptom identification.

\bibliographystyle{coling}
\bibliography{coling2020}

\end{document}